\documentclass[runningheads]{llncs}

\usepackage{eccvabbrv}

\usepackage{graphicx}
\usepackage{booktabs}

\usepackage[accsupp]{axessibility}  
\usepackage{multirow}
\usepackage{algorithm}
\usepackage{algorithmic}

\usepackage{hyperref}

\usepackage{orcidlink}

\begin{document}
\title{\textbf{Q$^2$}: Quantization-Aware Gradient Balancing and Attention Alignment for Low-Bit Quantization} 

\titlerunning{Q$^2$ for Low-Bit Quantization}

\author{First Author\inst{1}\orcidlink{0000-1111-2222-3333} \and
Second Author\inst{2,3}\orcidlink{1111-2222-3333-4444} \and
Third Author\inst{3}\orcidlink{2222--3333-4444-5555}}

\author{Zhaoyang Wang\inst{1} \and Dong Wang\inst{1}}

\authorrunning{Z.~Wang and D.~Wang}

\institute{Institute of Information Science, Beijing Jiaotong University,\\
No.3 Shangyuan Village, Xizhimenwai, Haidian District, Beijing, China, 100044\\
\email{24125200@bjtu.edu.cn \and wangdong@bjtu.edu.cn}}


\maketitle

\begin{abstract}
Quantization-aware training (QAT) has achieved remarkable success in low-bit ($\leq$4-bit) quantization for classification networks. However, when applied to more complex visual tasks such as object detection and image segmentation, performance still suffers significant degradation. A key  cause of this limitation has been largely overlooked in the literature. In this work, we revisit this phenomenon from a new perspective and identify a major failure factor: gradient imbalance at feature fusion stages, induced by accumulated quantization errors. This imbalance biases the optimization trajectory and impedes convergence under low-bit quantization.
Based on this diagnosis, we propose Q$^2$, a two-pronged framework comprising: (1) Quantization-aware Gradient Balancing Fusion (Q-GBFusion), a closed-loop mechanism that dynamically rebalances gradient contributions during feature fusion; and (2) Quantization-aware Attention Distribution Alignment (Q-ADA), a parameter-free supervision strategy that reconstructs the supervision distribution using semantic relevance and quantization sensitivity, yielding more stable and reliable supervision to stabilize training and accelerate convergence.
Extensive experiments show that our method, as a plug-and-play and general strategy, can be integrated into various state-of-the-art QAT pipelines, achieving an average +2.5\% mAP gain on object detection and a +3.7\% mDICE improvement on image segmentation. Notably, it is applied only during training and introduces no inference-time overhead, making it highly practical for real-world deployment.
\end{abstract}

\section{Introduction}
\label{sec:intro}

Model quantization reduces memory footprint and computational cost by approximating full-precision values with low-bit integers, and has become a foundational technique for neural network compression\cite{zs}. Existing quantization methods mainly fall into Post-Training Quantization (PTQ) and Quantization-Aware Training (QAT). PTQ minimizes quantization error on a small calibration set. QAT trains with quantization in the loop (fine-tuning or from scratch) and usually yields higher accuracy, particularly at $\leq$4-bit. State-of-the-art QAT approaches have demonstrated remarkable performance; in some cases, quantized models even outperform their full-precision counterparts \cite{Liu2022Nonuniform-to-Uniform,huang2025hessian,2025scheduling}. For instance, the recent N2UQ method \cite{Liu2022Nonuniform-to-Uniform} achieves a Top-1 accuracy of 78\% on a 4-bit ResNet-50, surpassing the full-precision baseline by 1\%.



\begin{figure*}[!tp]
\centering
\begin{minipage}{0.40\linewidth}
\includegraphics[width=\linewidth]{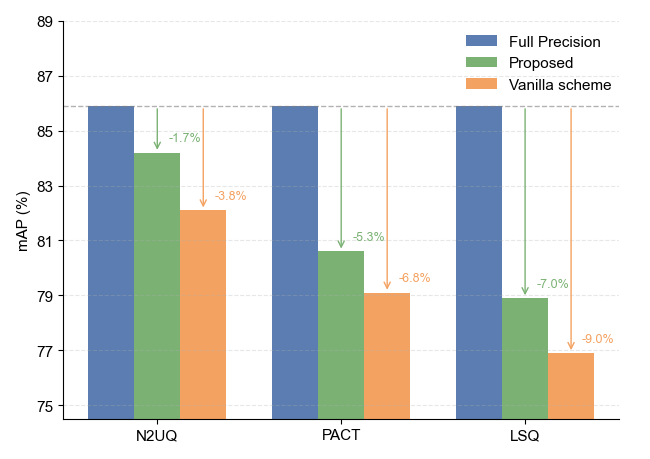}
\centerline{(a)}
\end{minipage}
\begin{minipage}{0.40\linewidth}
\includegraphics[width=\linewidth]{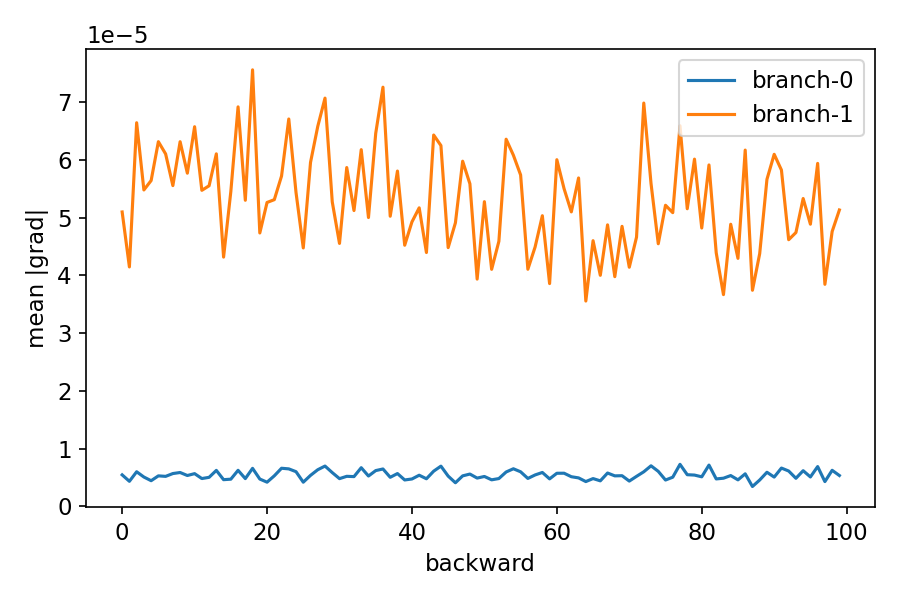}
\centerline{(b)}
\end{minipage}
\caption{(a) Comparison of the accuracy drop of three representative QAT schemes. YOLOv5 is used as the baseline model with  4-bit quantization setting. (b) Measured average gradient magnitude in the Concat Layer of the YOLOv5 model shown in Fig.~\ref{yolo}. To verify that this is a quantization-induced and general phenomenon, the corresponding measurements on other models are reported in the \textbf{Appendix}~\ref{Imbalance}.}
\label{fig}
\end{figure*}

However, applying existing QAT methods at 4-bit precision to more complex vision tasks such as object detection and image segmentation remains highly challenging. Only a limited number of related studies have been reported in the literature\cite{EMA,HQOD,qyolo}, and the achieved performance still falls short of that attained by quantized classification models. 
To better illustrate the performance gap, we apply three mostly cited schemes, including PACT\cite{pact}, LSQ\cite{LSQ} and N2UQ\cite{Liu2022Nonuniform-to-Uniform}, to the YOLO and compare the accuracy of the quantized model in Fig.~\ref{fig}(a). It is clear that even with the most powerful non-uniform quantization method N2UQ, the quantized model still suffers a 3.8\% accuracy loss.

These observations suggest that a quantizer-centric explanation is insufficient to account for low-bit degradation in complex visual tasks. Although models such as ResNet and YOLO are quantized with the same convolutional operator in implementation, their performance drops differ substantially under low-bit settings. This discrepancy indicates that quantization efficacy is shaped not only by the quantizer itself, but also by architectural characteristics. Existing studies mainly focus on quantization representations themselves: either by redesigning network architectures to improve quantization-friendliness~\cite{reg-ptq,chu2024make}, or by further optimizing quantizers ~\cite{EMA,qdetr,aqdetr}, while implicitly assuming that the optimization path itself remains reliable.
However, in detection and segmentation networks with feature fusion structures, this assumption does not always hold.



\begin{figure}
    \centering
    \includegraphics[width=0.7\linewidth]{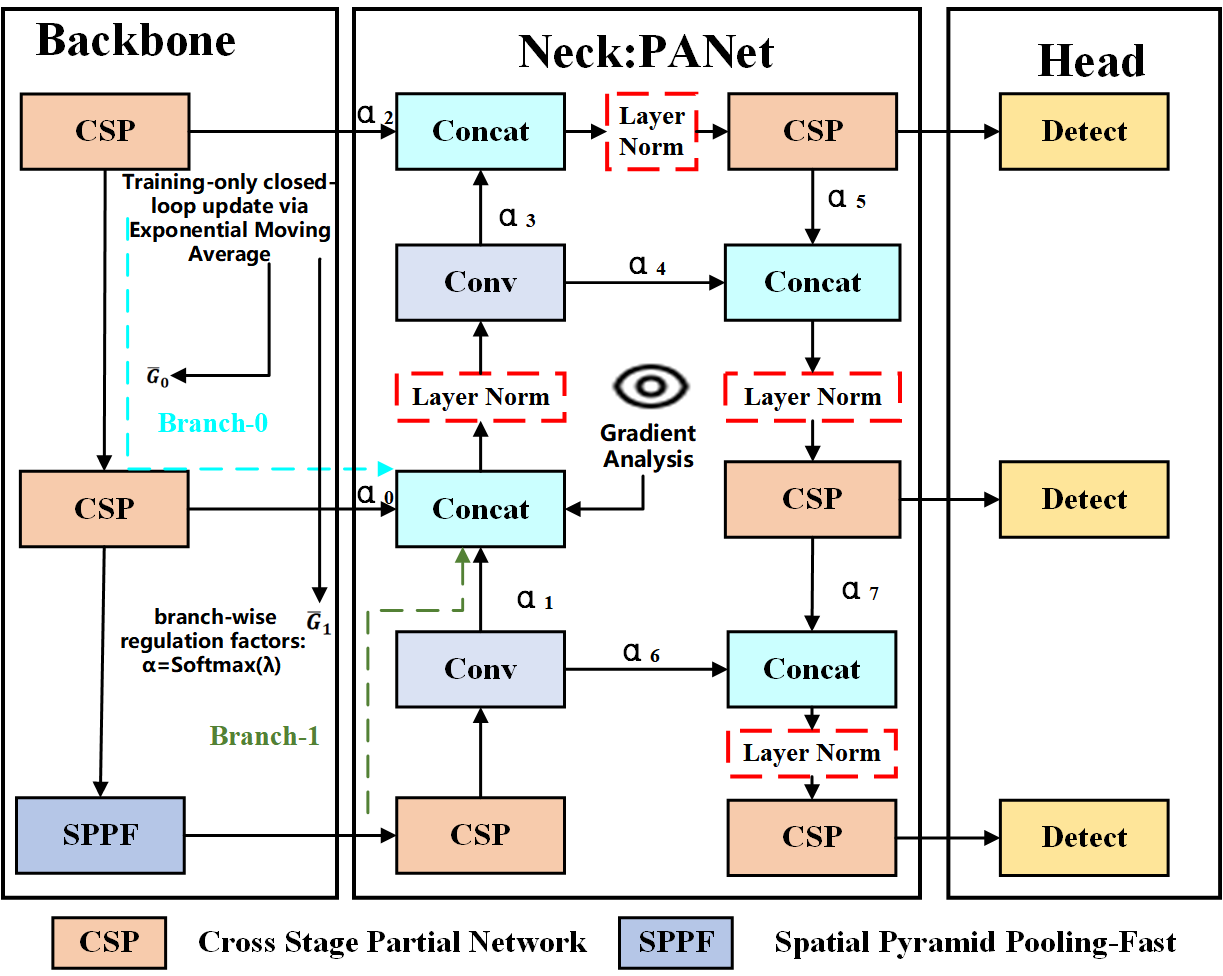}
    \caption{Applying the proposed feature fusion strategy to the YOLO network.  ${\alpha_i}$ denotes the \emph{regulation factors} introduced by Q-GBFusion.}
    \label{yolo}
\end{figure}

In this work, we revisit this problem from the perspective of optimization dynamics at feature-fusion stages. Taking YOLO as a representative detection architecture, as shown in Fig.~\ref{yolo}, its backbone and neck jointly support localization and recognition through multi-scale feature fusion~\cite{PAN}. In this process, shallow features (Branch-0) preserve fine-grained spatial details~\cite{maskrcnn}, whereas deep features (Branch-1) carry more abstract semantic information~\cite{TONG2024107931}. Unlike classification networks that mainly rely on final-layer high-level features, detection and segmentation models depend heavily on this multi-scale fusion mechanism for precise prediction.

We observe that under low-bit QAT, quantization errors progressively accumulate with network depth, leading to mismatched quantization-induced perturbation strength across different branches. When these branches are fused at fusion nodes, the backpropagated signal exhibits clear gradient imbalance. To quantify this phenomenon, based on the feature-gradient principle of Grad-CAM~\cite{gradcam}, we perform a feature gradient-flow analysis at the Concat layer (Fig.~\ref{yolo}) by tracking the average gradient magnitude. The results are shown in Fig.~\ref{fig}(b). We find a significant discrepancy between the two branches: QAT tends to disproportionately prioritize deeper branches (Branch-1) while relatively under-optimizing shallower ones (Branch-0), ultimately causing biased gradient updates at fusion points and degraded quantization performance. This effect becomes especially pronounced under ultra-low-bit settings (e.g., $\leq$4-bit).

Notably, the branch-wise gradient imbalance is non-stationary: its magnitude varies across fusion-layer locations, training stages, and quantization perturbations. Therefore, a fixed balancing coefficient is prone to under- or over-compensation during training, motivating an online feedback-based closed-loop regulation. To address this, we propose \textbf{Quantization-aware Gradient Balancing Fusion (Q-GBFusion)}, which introduces branch-wise regulation factors $\alpha_i$ (Fig.~\ref{yolo}) at fusion nodes and performs closed-loop adjustment based on gradient-energy feedback $G_i$ to balance the optimization of different branches, while applying post-fusion normalization to stabilize gradient propagation under low-bit quantization noise. During inference deployment, the closed-loop feedback update is disabled, and the related modules can be folded into parameters, thus introducing no additional runtime overhead.

Moreover, this imbalance is often accompanied by unstable QAT optimization and slower convergence to well-calibrated quantization parameters, since conventional QAT losses primarily minimize numerical discrepancies to ground-truth targets. However, for tasks such as object localization, predictions rely heavily on fine-grained semantic cues (e.g., shape, texture, and other feature representations), rather than numerical fidelity alone. Unlike existing methods that directly match feature tensors to enforce feature fidelity~\cite{Zhu_He_Wu_2023,Wei_2018_ECCV,boo2021stochastic}, our goal is to preserve not only semantic saliency but also quantization sensitivity; otherwise, distillation may amplify unreliable regions. Based on this analysis, we propose a quantization-distortion-aware attention distillation objective, termed \textbf{Quantization-aware Attention Distribution Alignment (Q-ADA)}. It aligns the full-precision teacher and quantized student by matching quantization-sensitive saliency-feature distributions via a saliency statistics rule, with greater emphasis on distortion-prone regions. This encourages the student to preserve fine-grained structural cues critical for downstream tasks, improving stability and accelerating convergence. Our contributions are summarized as follows:



\begin{itemize}
    \item \textbf{Mechanism-driven diagnosis.}
    We provide the first in-depth analysis showing that the performance degradation of low-bit quantization on complex vision tasks arises from a previously underexplored optimization pathology at feature-fusion stages, namely \emph{branch-wise gradient imbalance} caused by accumulated quantization errors, which further biases training optimization and leads to slow convergence.
    \item \textbf{Methodological contribution.}
    Guided by the above diagnosis, we propose \textbf{Q$^2$}, a two-pronged framework consisting of two complementary components: \textbf{Q-GBFusion}, which performs online feedback control of branch gradient allocation at feature-fusion stages to balance gradients, and \textbf{Q-ADA}, which enforces quantization-aware alignment of salient feature distributions to accelerate convergence.
    Q$^2$ is training-time only and introduces no extra inference cost.
    \item \textbf{Empirical contribution.}
    Extensive experiments across different architectures (CNNs and Transformers), tasks (object detection and image segmentation), and QAT pipelines show that the proposed method can be consistently integrated into diverse QAT methods with sustained performance gains, achieving an average +2.5\% mAP improvement on object detection and a +3.7\% mDICE improvement on image segmentation.
\end{itemize}

\section{Related Work}
\label{related}
\subsection{Quantization-Aware Training}
Existing QAT approaches can be broadly categorized into two complementary research directions. The first focuses on quantizer design, aiming to improve the fidelity of the quantization mapping itself, i.e., how floating-point weights and activations are optimally converted to low-bit fixed-point representations. Representative efforts include learnable scaling factors (e.g., LSQ\cite{LSQ}), adaptive clipping ranges (e.g., PACT\cite{pact}), non-uniform quantization schemes (e.g., N2UQ\cite{Liu2022Nonuniform-to-Uniform}), and compress–expand asymmetric quantization methods (e.g., LCQ\cite{LCQ}).
These methods seek to minimize quantization-induced distortion at the operator level. The second line of work addresses the optimization dynamics of QAT, recognizing that even with an accurate quantizer, standard training may converge to suboptimal solutions due to gradient mismatch, representation collapse, or loss of task-relevant features. To mitigate this, researchers have introduced enhanced supervision strategies such as transfer-rate scheduling and adaptive learning rate adjustment to improve QAT stability \cite{2025scheduling}, as well as delayed QAT activation with bit-width–controlled regularization for domain generalization \cite{qtdog}.
These approaches treat quantization not merely as a numerical approximation problem but as a representation learning challenge. 

\subsection{Quantization of Complex Visual Tasks}
Unlike traditional classification tasks, object detection and image segmentation rely on more complex network architectures that can achieve near full-precision performance at 8-bit or higher precision \cite{qyolo}, yet still suffer significant degradation at 4-bit (or lower) precision. Recent studies have attempted to narrow this gap. For example, HQOD \cite{HQOD} introduces task-correlated losses to balance regression optimization across IoU levels; however, it does not fully close the gap, and substantial degradation remains under 4-bit quantization (e.g., nearly a 7\% mAP drop on YOLO). Some methods further design quantization strategies for specific architectures (e.g., YOLO or DETR) \cite{EMA,qdetr,aqdetr}. While effective in targeted settings, they still leave a substantial gap under very low-bit quantization and have limited general applicability. This indicates that existing methods have not explicitly identified the shared structural bottleneck in complex vision networks, namely feature fusion stages. Some works recognize that architecture itself plays a critical role in quantization robustness and improve compatibility through architectural modifications \cite{reg-ptq,chu2024make}. For example, \cite{reg-ptq} replaces YOLO’s CSP backbone with a more “quantization-friendly” ResNet variant. However, such changes often compromise YOLO’s original design principles, leading to degraded baseline performance and reduced practical utility.

These observations suggest the need for a more general, architecture-agnostic solution that targets the shared structural bottleneck in complex vision networks—namely, feature fusion stages—while preserving inference efficiency.

\section{Proposed Method}
\subsection{Problem Formulation}
\label{sec:formulation}

During quantization-aware training, each layer introduces quantization error.
Let $x_l$ denote the full-precision output of the $l$-th layer and $\tilde{x}_l$ its quantized counterpart, with $\delta_l=\tilde{x}_l-x_l$.
Since the input to layer $l$ is also quantized, $\tilde{x}_{l-1}=x_{l-1}+\delta_{l-1}$, the quantized output can be written as
$\tilde{x}_l = f_l(\tilde{x}_{l-1}) + \epsilon_l \approx f_l(x_{l-1}+\delta_{l-1})+\epsilon_l$,
where $f_l(\cdot)$ is the full-precision transformation of the $l$-th layer and $\epsilon_l$ is the layer-wise quantization noise.
Under a first-order approximation,
\begin{equation}
\delta_l \approx J_l \delta_{l-1} + \epsilon_l,
\label{eq:err_rec}
\end{equation}
where $J_l=\partial f_l/\partial x_{l-1}$ is the Jacobian evaluated at $x_{l-1}$.
Eq.~(\ref{eq:err_rec}) shows that quantization disturbances propagate and accumulate with depth, making deep representations more sensitive under low-bit settings.

Modern detectors/segmentors rely on multi-scale feature fusion across multiple branches.
Consider a fusion node with $K$ quantized branch features $\{\tilde{\mathbf F}_i\}_{i=1}^{K}$ (e.g., concatenation), where $\tilde{\mathbf F}_i$ denotes the vectorized feature of the $i$-th branch and
$\tilde{\mathbf F}_i=\mathbf F_i+\delta_i$,
with $\mathbf F_i$ the corresponding full-precision feature and $\delta_i$ the feature-level quantization error.
Because different branches generally have different effective depths and receptive fields, the magnitudes of $\delta_i$ can vary substantially under low-bit quantization, and the disparity becomes more pronounced as bit-width decreases.

To analyze how quantization affects localization, we adopt a local linear proxy of the regression head around the fusion bottleneck:
\begin{equation}
\hat{\mathbf t} = \sum_{i=1}^{K} \mathbf W_i \tilde{\mathbf F}_i
= \sum_{i=1}^{K} \mathbf W_i (\mathbf F_i+\delta_i),
\label{eq:lin_pred}
\end{equation}
where $\hat{\mathbf t}\in\mathbb{R}^{m}$ denotes the localization output vector, $\mathbf t^{\*}\in\mathbb{R}^{m}$ is the corresponding ground-truth target, and $\mathbf W_i\in\mathbb{R}^{m\times n_i}$ is the effective regression weight of the $i$-th branch with $\tilde{\mathbf F}_i\in\mathbb{R}^{n_i}$.
For CIoU-style~\cite{ciou} regression, we approximate the loss sensitivity by squared error:
$\mathcal L_{\text{CIoU}} \propto \|\hat{\mathbf t}-\mathbf t^{\*}\|_2^2,$
which expands as
\begin{equation}
\hat{\mathbf t}-\mathbf t^{\*}
=
\sum_{i=1}^{K} \mathbf W_i \delta_i
+
\sum_{i=1}^{K} \mathbf W_i \mathbf F_i
-\mathbf t^{\*}.
\label{eq:err_expand}
\end{equation}

A key observation in low-bit QAT is that, due to the depth-wise accumulation in Eq.~(\ref{eq:err_rec}), some branches can incur larger effective disturbances $\mathbf W_i\delta_i$ at fusion bottlenecks, yielding biased backpropagated gradient flow.
To make Eq.~(\ref{eq:err_expand}) actionable for optimization, we summarize each branch by its gradient energy
$G_i \triangleq \big\|\partial\mathcal{L}/\partial \tilde{\mathbf F}_i\big\|_2$.
When $\mathbf W_i\delta_i$ dominates Eq.~(\ref{eq:err_expand}) for a subset of branches, their $G_i$ becomes disproportionately large, causing preferential optimization of those branches while others are under-updated; we therefore impose a multi-branch log-energy balance constraint.

Let
\(
\mathbf g_i \!=\! \partial \mathcal L/\partial \tilde{\mathbf F}_i
\)
denote the per-mini-batch gradient of the $i$-th branch at a fusion node.
We measure branch gradient energy by $\ell_2$ norms and enforce log-domain balancing across all $K$ branches:
\begin{equation}
\mathbb E_{\mathcal B}\!\left[\log\left(\|\mathbf g_i\|_2+\epsilon\right)\right]
-
\frac{1}{K}\sum_{j=1}^{K}\mathbb E_{\mathcal B}\!\left[\log\left(\|\mathbf g_j\|_2+\epsilon\right)\right]
=\tau_i,\quad \forall i\in\{1,\ldots,K\},
\label{eq:grad_constraint}
\end{equation}
where $\mathbb E_{\mathcal B}[\cdot]$ denotes expectation over mini-batches (approximated online during training), $\epsilon>0$ is a numerical-stability constant inside the logarithm, and $\tau_i$ is a target offset (default $\tau_i=0$ for all $i$ to encourage uniform balancing).\footnote{Equivalently, Eq.~(\ref{eq:grad_constraint}) constrains the deviations of per-branch log-energies from their mean; setting all $\tau_i=0$ minimizes inter-branch imbalance in the log-energy domain.}
In this work, our goal is to design a \emph{plug-and-play} fusion mechanism that enforces Eq.~(\ref{eq:grad_constraint}) online during QAT without altering the original network topology or incurring inference overhead.

\subsection{Quantization-Aware Gradient Balancing Fusion}
\label{sec:qgbfusion}
As analyzed above, branch-wise gradient imbalance at fusion bottlenecks under low-bit QAT is inherently non-stationary, varying across layers, training stages, and quantization perturbations. Therefore, gradient balancing should be treated as a dynamic optimization problem rather than addressed by a fixed coefficient. A static allocation may violate the desired gradient-energy constraint over time, leading to biased updates. This motivates a feedback-driven regulation mechanism that adaptively enforces the multi-branch gradient-energy constraint throughout training. Empirical evidence with fixed coefficients is provided in the \textbf{Appendix}~\ref{addition}. We propose \textbf{Quantization-aware Gradient Balancing Fusion (Q-GBFusion)}, a plug-and-play module that enforces the multi-branch gradient-energy constraint in Eq.~(\ref{eq:grad_constraint}) at fusion bottlenecks.
Q-GBFusion is \emph{training-only controllable}: it requires no second-order gradients, does not modify the backbone topology, and introduces no inference-time overhead.


Consider a fusion node with $K$ quantized branch features $\{\tilde{\mathbf F}_i\}_{i=1}^{K}$.
We maintain an unconstrained internal state (dual logits) $\boldsymbol{\lambda}\in\mathbb R^{K}$ and obtain a simplex-constrained allocation vector by a softmax projection:
\begin{equation}
\boldsymbol{\alpha}=\mathrm{Softmax}(\boldsymbol{\lambda}),\qquad
\mathbf F_i'=\alpha_i\cdot \tilde{\mathbf F}_i,\ \ i=1,\ldots,K,
\label{eq:dual_gate_multi}
\end{equation}
followed by the original fusion operator, e.g., $\mathbf F_{\text{cat}}'=[\mathbf F_1';\ldots;\mathbf F_K']$.
This parameterization ensures $\alpha_i\ge 0$ and $\sum_i\alpha_i=1$, and admits a closed-loop interpretation: $\boldsymbol{\lambda}$ acts as a dual control state whose projection determines the fusion allocation among branches.

Immediately after fusion, we insert a LayerNorm~\cite{2016Layer} module applied across channels at each spatial location:
\begin{equation}
\mathrm{LN}(h)=\frac{h-\mu_h}{\sigma_h},\qquad h=\mathbf F_{\text{fuse}}'(i,j),
\label{eq:ln_forward}
\end{equation}
where $\mathbf F_{\text{fuse}}'$ denotes the fused feature (e.g., $\mathbf F_{\text{cat}}'$), $(i,j)$ indexes a spatial location, and $\mu_h,\sigma_h$ are computed over the channel dimension (per sample and per location).

\subsubsection{Training-time: closed-loop gradient-energy balancing.}
\label{sec:qgbfusion_train}
Let $\mathbf g_i=\partial\mathcal L/\partial \tilde{\mathbf F}_i$ denote the per-mini-batch branch gradient at this fusion node, and define the corresponding gradient energy $G_i=\|\mathbf g_i\|_2$.
By the chain rule applied to Eq.~(\ref{eq:dual_gate_multi}), we have $\mathbf g_i=\alpha_i\,\partial\mathcal L/\partial \mathbf F_i'$.
Taking log-norms yields a per-branch decomposition
\begin{equation}
\log(G_i+\epsilon)=\log(\alpha_i+\epsilon)+\log\!\left(\left\|\frac{\partial\mathcal L}{\partial \mathbf F_i'}\right\|_2+\epsilon\right),
\label{eq:log_decomp_multi}
\end{equation}
which separates the controllable allocation term $\log(\alpha_i+\epsilon)$ from a residual term that captures gradient statistics after gating and post-fusion conditioning.
Therefore, balancing branch energies in the log domain can be achieved by regulating $\boldsymbol{\alpha}$ (hence $\boldsymbol{\lambda}$) using gradient-energy feedback.

In practice, we estimate the expectations in Eq.~(\ref{eq:grad_constraint}) using Exponential Moving Average (EMA) statistics $\{\bar G_i\}_{i=1}^{K}$ of $\{G_i\}_{i=1}^{K}$:
\begin{equation}
\bar G_i \leftarrow (1-\beta)\bar G_i + \beta\, G_i,\qquad i=1,\ldots,K,
\label{eq:ema_multi}
\end{equation}
where $\beta\in(0,1)$ is the EMA momentum.
We then form log-energy deviations from the mean:
\begin{equation}
e_i = \log(\bar G_i+\epsilon) - \frac{1}{K}\sum_{j=1}^{K}\log(\bar G_j+\epsilon) - \tau_i,\qquad i=1,\ldots,K,
\label{eq:error_multi}
\end{equation}
where $\tau_i$ is a target offset (default $\tau_i=0$ for uniform balancing).
Finally, we update the dual logits by a simple first-order feedback law:
\begin{equation}
\lambda_i \leftarrow \lambda_i - \eta\, e_i,\qquad i=1,\ldots,K,
\label{eq:lambda_update_multi}
\end{equation}
with step size $\eta>0$.

During training, the network weights are optimized by standard backpropagation, while the fusion allocation is adjusted by gradient-energy feedback to reduce biased updates across branches.
After training, the learned allocation $\boldsymbol{\alpha}$ can be fixed, and the closed-loop update in Eqs.~(\ref{eq:ema_multi})--(\ref{eq:lambda_update_multi}) is disabled,
hence introducing no extra operators or inference overhead.

\subsubsection{Deployment-time: LayerNorm-free inference.}
\label{sec:qgbfusion_deploy}

LayerNorm (LN) is generally viewed as ``non-removable'' because it is input-dependent: the statistics $\mu_h$ and $\sigma_h$ vary with $h$, so replacing them by fixed constants can cause a noticeable mismatch.
However, in low-bit quantization the feature space is severely range-limited, making the per-sample LN statistics much less variant.
In this work, we propose a "removable" approach: compute calibration statistics on a small held-out set and approximate LN by a fixed affine transform:
\begin{equation}
\mathrm{LN}(h)=\frac{h-\mu_h}{\sigma_h}\ \approx\ \frac{h-\bar\mu}{\bar\sigma},
\label{eq:ln_affine_approx}
\end{equation}
where $(\bar\mu,\bar\sigma)$ are calibration estimates of $(\mu_h,\sigma_h)$.

Let a downstream linear map be $y=\mathbf W h_{\text{LN}}+\mathbf b$ with $h_{\text{LN}}=\mathrm{LN}(h)$.
Using Eq.~(\ref{eq:ln_affine_approx}), we obtain an equivalent re-parameterization
\begin{equation}
y \approx \mathbf W' h + \mathbf b',\qquad
\mathbf W'=\frac{1}{\bar\sigma}\mathbf W,\quad
\mathbf b'=\mathbf b-\frac{\bar\mu}{\bar\sigma}\mathbf W\mathbf 1,
\label{eq:ln_fold}
\end{equation}
where $\mathbf 1$ is an all-ones vector with compatible dimension (capturing that $\bar\mu$ is subtracted uniformly across channels at each spatial location).
Thus, LN can be safely removed at deployment by (i) calibrating $(\bar\mu,\bar\sigma)$ and (ii) folding Eq.~(\ref{eq:ln_fold}) into the following layer.
Detailed LayerNorm-removal derivations for Eq.~(\ref{eq:ln_affine_approx}) are provided in the \textbf{Appendix} \ref{LayerNormremove}, and empirical accuracy validations are reported in Sec.~\ref{ablation1}.

\subsection{Quantization-Aware Attention Distribution Alignment}


Conventional quantization methods typically supervise training end-to-end using task-level losses against ground truth (e.g., classification or regression errors), yet this paradigm often neglects the fidelity of intermediate feature representations. This limitation becomes especially pronounced for regression-heavy objectives such as bounding-box localization, where quantization errors can be amplified because accurate predictions rely on fine-grained spatial cues. Motivated by this, we argue that salient feature information should be explicitly incorporated to guide QAT in low-bit regimes.

Several feature-level supervision schemes have been proposed in the literature, such as ~\cite{Zhu_He_Wu_2023}\cite{Wei_2018_ECCV}\cite{boo2021stochastic}. However, they tend to be fragile under QAT: most of these methods directly match intermediate feature tensors, while quantization introduces non-stationary perturbations whose magnitude and pattern evolve during training (e.g., due to changing step sizes, clipping ranges, and rounding behavior). Consequently, the supervision target itself can drift, making strict tensor-level alignment unstable and preventing the student from reliably preserving fine-grained information.



To address this, we propose \textbf{Quantization-aware Attention Distribution Alignment (Q-ADA)}, a distribution-level distillation scheme that is explicitly designed for low-bit QAT. Rather than enforcing point-wise feature matching, Q-ADA aligns attention distributions using quantization-aware statistics that are more stable under evolving quantization noise: (i) \textit{mean-centered responses} (deviation from the per-channel mean) to highlight saliency, (ii) \textit{channel-wise variance} to normalize dynamic range and accommodate activation scaling, and (iii) a \textit{local quantization distortion map} to emphasize regions/channels that are most vulnerable to quantization error. 

Given a feature map $X \in \mathbb{R}^{C\times H\times W}$, let $\mu_c$ and $\sigma_c^2$ denote the mean and variance of channel $c$ over spatial positions. During QAT, we denote the quantized feature map as $\widehat{X}=Q(X)$ and define the per-position quantization error
\begin{equation}
\Delta_{c,ij} \triangleq \big|\,X_{c,ij}-\widehat{X}_{c,ij}\,\big|.
\end{equation}
To ensure comparability across channels and bit-widths, we normalize the distortion by the per-channel quantization step size $s_c$ (or by $\sigma_c$ when $s_c$ is unavailable), and define two parameter-free statistics:
\begin{equation}
z_{c,ij} \triangleq \frac{|X_{c,ij}-\mu_c|}{\sigma_c+\kappa},
\qquad
r_{c,ij} \triangleq \frac{\Delta_{c,ij}}{s_c+\kappa},
\label{eq:qada_stats}
\end{equation}
where $\kappa>0$ is a small constant for numerical stability.

Finally, the overall \textit{saliency score} is calculated as:
\begin{equation}
S_{c,ij}(X)
\triangleq
\log\!\big(1+z_{c,ij}^2\big)
+\gamma\,\log\!\big(1+r_{c,ij}^2\big),
\label{eq:qada_energy}
\end{equation}
where $\gamma$ controls the strength of quantization-vulnerability emphasis. The first term highlights statistically salient locations relative to the channel distribution, while the second term explicitly increases the score for locations that incur larger quantization distortion.

The corresponding quantization-aware attention weight is computed as $\widetilde{A}_{c,ij}= \mathrm{Sigmoid}\!\big(S_{c,ij}(X)\big)
\label{eq:qada_attn} $. During distillation, we construct spatial probability distributions from the teacher (full-precision) and student (quantized) attention maps via normalization: $P_{c,ij}=\frac{\widetilde{A}^{\,f}_{c,ij}}
{\sum_{i',j'}\widetilde{A}^{\,f}_{c,i'j'}},
\qquad
R_{c,ij}=\frac{\widetilde{A}^{\,q}_{c,ij}}
{\sum_{i',j'}\widetilde{A}^{\,q}_{c,i'j'}},$ and align these distributions using the Jensen--Shannon divergence:
\begin{equation}
\mathcal{L}^{(c)}_{\text{ADA}}
= \tfrac{1}{2}\!\sum_{i,j} P_{c,ij}\log\frac{P_{c,ij}}{M_{c,ij}}
+ \tfrac{1}{2}\!\sum_{i,j} R_{c,ij}\log\frac{R_{c,ij}}{M_{c,ij}},
\label{eq:qada_js}
\end{equation}
where $M_c=\tfrac{1}{2}\,(P_c+R_c)$.
Notably, the divergence choice is not unique; Section~\ref{ablation1} compares KL and its corresponding impact on performance. A visualization of the Q-ADA distillation process is provided in \textbf{Appendix}~\ref{Visualizing}.


\section{Experiments}

\subsection{Experimental Setup}
We evaluate the proposed method on two representative visual tasks: object detection and image segmentation.
For object detection, we adopt the latest CNN-based YOLOv11 and the widely cited YOLOv5 models, and further evaluate our method on the transformer-based RT-DETR\cite{rtdetr}.
Experimental results are reported on the PASCAL VOC \cite{VOC} and COCO \cite{coco} datasets using mean Average Precision (mAP) as the primary evaluation metric.
For image segmentation, we employ MK-UNet \cite{Rahman_2025_ICCV}, a recent state-of-the-art architecture. Experiments are conducted on the BUSI medical imaging dataset \cite{BUSI}, with performance measured by the mean Dice coefficient (mDICE).
All experiments are performed on a server equipped with 8 $\times$ NVIDIA GeForce RTX 4090 GPUs. More detailed implementations and settings are provided in the \textbf{Appendix} \ref{Setup}.




\subsection{Evaluation of General Effectiveness }
\label{sss}
First, we integrate Q-GBFusion and Q-ADA with different QAT quantizers. Specifically, PACT, LSQ, and N2UQ are adopted as convolution-oriented quantizers, while Q-DETR\cite{qdetr}, AQ-DETR\cite{aqdetr}, and GPLQ\cite{gplq} are employed to quantize attention operators. 
We only report results on the VOC dataset in the main text, and provide COCO results in the \textbf{Appendix} \ref{Supplementary Results}.


\begin{table}[htbp]
\centering
\caption{Quantization results of the object-detection models on the VOC dataset.}
\label{tab:integration_models_bitwidths}
\scriptsize
\setlength{\tabcolsep}{1pt}
\renewcommand{\arraystretch}{0.35}

\resizebox{0.75\linewidth}{!}{
\begin{tabular}{c c c c c c c}
\toprule
\textbf{Networks} & \textbf{Structure} & \textbf{BW} & \textbf{Method} & \textbf{Baseline} & \textbf{+ Ours} & \textbf{Gain}\\
\midrule
\multirow{6}{*}{\begin{tabular}[c]{@{}c@{}}YOLOv5s\\ (FP: 85.9\%)\end{tabular}}
& \multirow{6}{*}{CNN}
& \multirow{3}{*}{W4A4}
& N2UQ  & 82.1\% & \textbf{84.2\%} & +2.1 \\
& & & PACT  & 79.1\% & \textbf{80.6\%} & +1.5 \\
& & & LSQ   & 76.9\% & \textbf{78.9\%} & +2.0 \\
\cmidrule(lr){3-7}
& & \multirow{3}{*}{W3A3}
& N2UQ  & 75.5\% & \textbf{78.0\%} & +2.5 \\
& & & PACT  & 62.9\% & \textbf{66.6\%} & +3.7 \\
& & & LSQ   & 59.9\% & \textbf{66.8\%} & +6.9 \\
\midrule
\multirow{6}{*}{\begin{tabular}[c]{@{}c@{}}YOLOv11s\\ (FP: 89.4\%)\end{tabular}}
& \multirow{6}{*}{CNN}
& \multirow{3}{*}{W4A4}
& N2UQ  & 86.2\% & \textbf{87.6\%}& +1.4 \\
& & & PACT  & 83.8\% & \textbf{84.8\%} & +1.0 \\
& & & LSQ   & 82.9\% & \textbf{84.2\%} & +1.3 \\
\cmidrule(lr){3-7}
& & \multirow{3}{*}{W3A3}
& N2UQ  & 83.0\% & \textbf{84.3\%} & +1.3 \\
& & & PACT  & 75.1\% & \textbf{78.2\%} & +3.1 \\
& & & LSQ   & 75.5\% & \textbf{78.9\%} & +3.4 \\
\midrule
\multirow{6}{*}{\begin{tabular}[c]{@{}c@{}}RT-DETR\\ (FP: 90.1\%)\end{tabular}}
& \multirow{6}{*}{Transformer}
& \multirow{3}{*}{W4A4}
& Q-DETR   & 80.2\% & \textbf{82.4\%}& +2.2 \\
& & & AQ-DETR  & 81.4\% & \textbf{83.3\%} & +1.9 \\
& & & GPLQ     & 83.7\% & \textbf{86.3\%} & +2.6 \\
\cmidrule(lr){3-7}
& & \multirow{3}{*}{W3A3}
& Q-DETR   & 77.1\% & \textbf{79.8\%} & +2.7 \\
& & & AQ-DETR  & 77.5\% & \textbf{80.0\%} & +2.5 \\
& & & GPLQ     & 78.3\% & \textbf{81.8\%} & +3.5 \\
\bottomrule
\end{tabular}
}
\end{table}

The object detection results are summarized in Table~\ref{tab:integration_models_bitwidths}.
Our method consistently improves accuracy across all baseline quantizers and network architectures. Specifically, under different bit-width settings, it achieves an average gain of up to +2.5\% mAP, showing strong generalization regardless of the underlying quantization scheme. 
The gains become even more pronounced under the stricter W3A3 setting (up to +6.9\% improvement), where quantization noise significantly degrades feature representation quality. 
Notably, when combined with N2UQ, it further narrows the accuracy gap to within 2\% of the FP model, highlighting compatibility with advanced quantization designs.





We further validate our approach on an image segmentation model, MK-UNet, as reported in Table~\ref{tab2}. Across all quantizers and bit-widths, our method yields an average +3.7\% mDICE gain, with improvements reaching +4.9\% under W3A3. These results confirm that the proposed strategy not only benefits detection tasks but also generalizes well to segmentation problems, where fine-grained spatial consistency is critical.
Despite an 8.8\% mDice gap relative to the FP baseline, our method (W4A4) notably outperforms the current 8-bit SOTA quantization scheme\cite{ZHANG2024103277} by +4.4\%. 

\begin{table}[!htbp]
\centering
\caption{Quantization results of the segmentation model on the BUSI dataset.}
\label{tab2}
\small
\resizebox{0.6\linewidth}{!}{%
\begin{tabular}{c c c c c c}
\toprule
\textbf{Networks} & \textbf{BW} & \textbf{Method} & \textbf{Baseline} & \textbf{+ Ours} & \textbf{Gain} \\
\midrule
\multirow{7}{*}{\begin{tabular}[c]{@{}c@{}}MK-UNet\\ (FP: 69.5\%)\end{tabular}}
& \multirow{1}{*}{W8A8}
& EQ\cite{ZHANG2024103277} & 55.9\% & -- & -- \\
\cline{2-6}
& \multirow{3}{*}{W4A4}
& N2UQ  & 55.4\% & \textbf{60.7\%} & +5.3 \\
& & PACT  & 44.5\% & \textbf{46.3\%} & +1.8 \\
& & LSQ   & 45.6\% & \textbf{49.7\%} & +4.1 \\
\cmidrule(lr){2-6}
& \multirow{3}{*}{W3A3}
& N2UQ  & 46.5\% & \textbf{53.9\%} & +7.4 \\
& & PACT  & 39.3\% & \textbf{42.7\%} & +3.4 \\
& & LSQ   & 40.9\% & \textbf{45.4\%} & +4.5 \\
\bottomrule
\end{tabular}%
}
\end{table}

\subsection{Comparison with SOTA Optimization Methods} 
\label{ss1}
We further evaluate the proposed framework against other optimization strategies across different quantization baselines and tasks. Specifically, Table~\ref{tab:qat_comparison_combined} reports comparisons with state-of-the-art training-optimization-centric QAT methods. 
The reference results are rigorously re-implemented according to the original papers. More details on the reproduction procedure and configuration setup are provided in the \textbf{Appendix} \ref{Reproduction}. 

Compared to the baseline, most reference optimization approaches yield only marginal improvements. Since EMA is specifically designed for YOLO, it exhibits obvious advantage over other reference methods. However, when transferred to MK-UNet, its accuracy drops drastically.
This indicates that existing strategies are insufficient for addressing the inherent gradient imbalance problem in feature fusion. In contrast, the proposed method substantially alleviates this issue, achieving an average improvement of 3\%–4\% over competing approaches.


Finally, we augment both QT-DoG and EMA
with our proposed approach. This yields consistent improvements of +4.6\% and +1.8\%, demonstrating that our method not only enhances the underlying quantizers but also provides complementary benefits to existing optimization strategies.

\begin{table}[t]
\centering
\caption{Performance comparison of different optimization strategies on YOLOv5 (COCO) and MK-UNet (BUSI). All results are under W4A4 quantization. The base quantizer is N2UQ for YOLOv5 and LSQ for MK-UNet.}
\label{tab:qat_comparison_combined}
\scriptsize
\setlength{\tabcolsep}{4pt}
\renewcommand{\arraystretch}{1.0}

\resizebox{0.98\linewidth}{!}{%
\begin{tabular}{lcc}
\toprule
\textbf{Strategy} & \textbf{YOLOv5 (COCO), mAP} & \textbf{MK-UNet (BUSI), mDICE} \\
\midrule
Baseline 
& 31.1\% \; (N2UQ) 
& 45.6\% \; (LSQ) \\

+ Compute-Optimal QAT \cite{dremov2026computeoptimal} (ICLR 2026)
& 31.7\% 
& 46.3\% \\

+ TR \cite{2025scheduling} (ICCV 2025)
& 31.4\% 
& 45.7\% \\

+ HMQAT \cite{huang2025hessian} (NN 2025)
& 31.3\% 
& 45.6\% \\

+ QT-DoG \cite{qtdog} (ICML 2025)
& 31.7\% 
& 46.8\% \\

+ EMA \cite{EMA} (WACV 2024)
& 32.2\% 
& 45.9\% \\
\midrule

+ Ours
& \textbf{33.2\%} \; (N2UQ + Ours)
& \textbf{49.7\%} \; (LSQ + Ours) \\
\midrule

+ Best combination
& \textbf{34.0\%} \; (N2UQ + EMA \cite{EMA} + Ours)
& \textbf{51.4\%} \; (LSQ + QT-DoG \cite{qtdog} + Ours) \\
\bottomrule
\end{tabular}%
}
\end{table}

\subsection{Visual Analysis}
\begin{figure*}[!tp]
\centering
\begin{minipage}{0.6\linewidth}
\includegraphics[width=\linewidth]{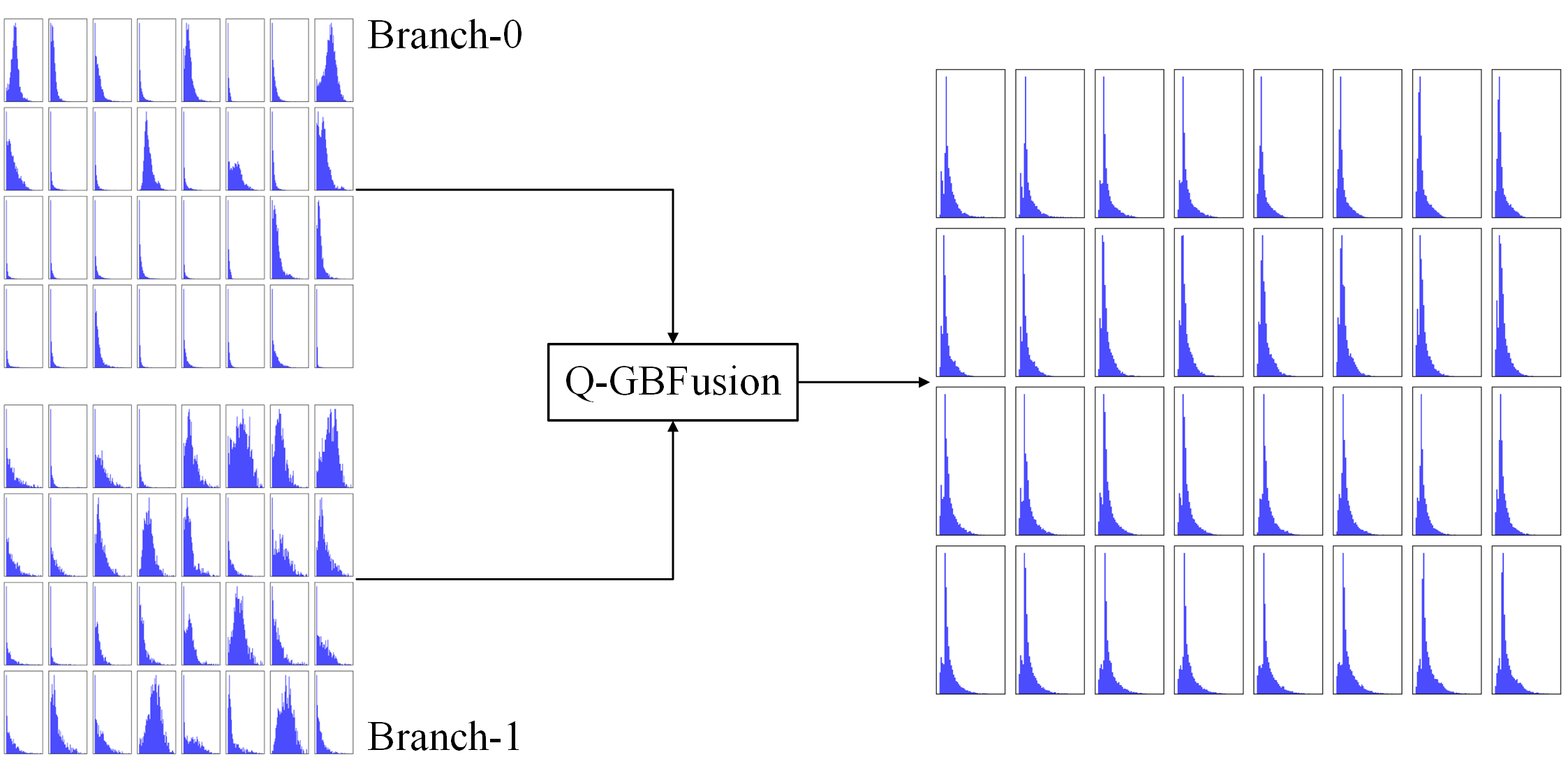}
\centerline{(a)}
\end{minipage}
\begin{minipage}{0.39\linewidth}
\includegraphics[width=\linewidth]{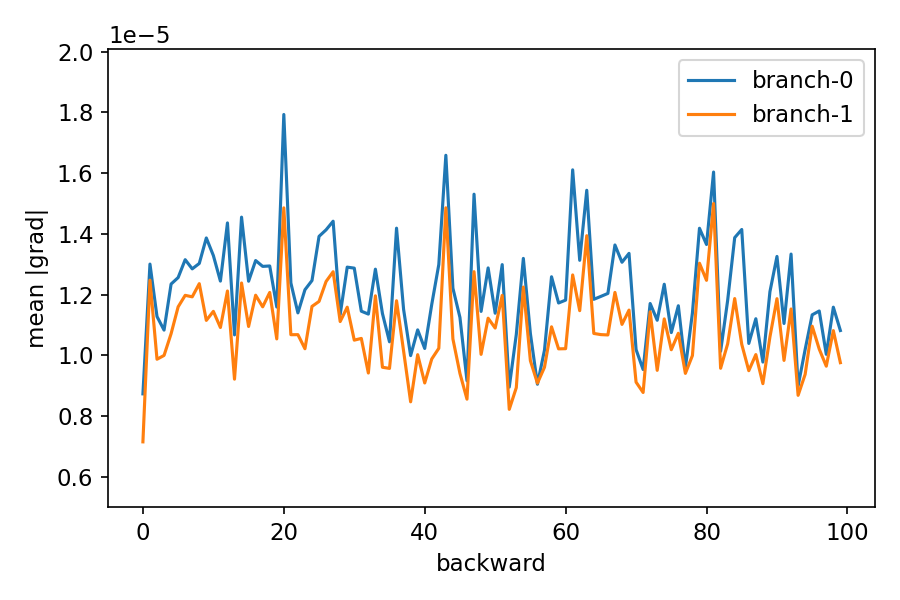}
\centerline{(b)}
\end{minipage}
\caption{(a) Visualization of feature magnitudes across 32 channels at the concatenation layer in YOLOv5. (b) Gradient magnitudes from the two branches at the feature fusion node in YOLOv5 after applying Q-GBFusion.}
\label{grad5}
\end{figure*}

To better illustrate how Q-GBFusion facilitates quantization-aware training
we provide, in this section, visualizations of intermediate representations during both forward and backward propagation in Fig.~\ref{grad5}.

In Fig.\ref{grad5}(a) , the original feature distributions exhibit significant disparities in dynamic range and scale. After applying Q-GBFusion, the fused outputs demonstrate a much more uniform and balanced distribution across all channels, indicating that our method effectively normalizes the cross-branch feature scales prior to fusion.

In the backward pass, Fig.\ref{grad5}(b) plots the mean gradient magnitudes flowing back through the two branches during quantization training. 
Compared to Fig.~\ref{fig}(b), it is clear that, after integrating Q-GBFusion, the gradient flows from both branches become significantly more balanced, with similar magnitude trajectories over training steps. 

\begin{figure}[!tp]
\centering
\begin{minipage}{0.45\linewidth}
\includegraphics[width=\linewidth]{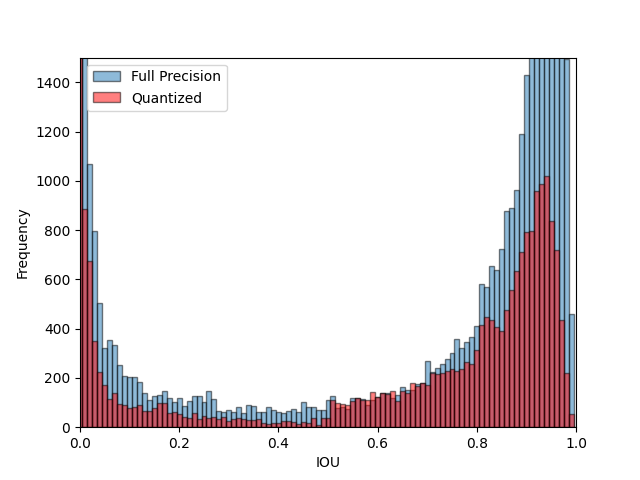}
\centerline{(a)}
\end{minipage}
\begin{minipage}{0.45\linewidth}
\includegraphics[width=\linewidth]{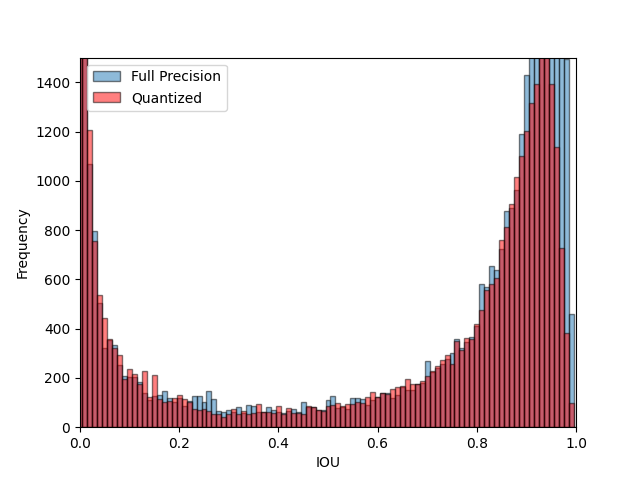}
\centerline{(b)}
\end{minipage}
\caption{Comparison of IoU distributions before and after integrating Q-ADA. (a) YOLOv11 quantized directly using N2UQ. (b) YOLOv11 quantized with Q-ADA integrated.}
\label{iou}
\end{figure}

To demonstrate how Q-ADA improves localization in quantized object detection, we compare the IoU distributions of YOLOv11 predictions before and after applying Q-ADA under 4-bit N2UQ quantization (Fig.~\ref{iou}). 
It is clear that, without Q-ADA, quantization shifts IoU downward, sharply reducing high-IoU boxes ($\geq$0.5/0.8) due to impaired spatial cues. Q-ADA aligns FP/quantized attention, preserving semantics and recovering accurate, high-confidence localization.

\begin{table}[!t]
\centering
\small
\caption{Component ablation on 4-bit quantized YOLOv5 (PASCAL VOC).}
\label{tab:abl_comp}
\setlength{\tabcolsep}{4.2pt}
\resizebox{0.75\linewidth}{!}{%
\begin{tabular}{l|cc|cc}
\toprule
\multirow{2}{*}{\textbf{Strategy}} & \multicolumn{2}{c|}{\textbf{N2UQ (W4A4)}} & \multicolumn{2}{c}{\textbf{LSQ (W4A4)}} \\
\cmidrule(lr){2-3} \cmidrule(lr){4-5}
& mAP & Time(h) & mAP & Time(h) \\
\midrule
Baseline & 82.1\% & 6.25 & 76.9\% & 3.80 \\
+ Q-GBFusion & 83.5\% & 6.47 & 78.4\% & 3.83 \\
+ Q-GBFusion + Q-ADA (KL) & 83.8\% & 3.31 & \textbf{78.9\%} & 2.01 \\
+ Q-GBFusion + Q-ADA (JS) & \textbf{84.2\%} & 2.98 & 78.5\% & 2.20 \\
\bottomrule
\end{tabular}%
}
\end{table}

\subsection{Ablation Study}
\label{ablation1}

\subsubsection{Component Ablation: Q-GBFusion vs. Q-ADA}
We report the component-wise ablation results in Table~\ref{tab:abl_comp}. \textit{Time} is measured under the same hardware/setup with validation-based early stopping, thus reporting time-to-convergence (not fixed epochs).  The teacher is a frozen pretrained full-precision model, which does not participate in quantized parameter updates and is used only to generate supervision signals for distillation..

It can be observed from the table that Q-GBFusion consistently improves mAP by 1.4--1.5\% across quantizers, indicating that balancing multi-scale fusion effectively stabilizes gradient propagation. Adding Q-ADA yields further gains (+0.3\% with KL and +0.7\% with JS) and substantially reduces training time due to faster convergence.


The choice of divergence interacts non-trivially with the quantization scheme.
KL divergence, which places greater emphasis on discrepancies in low-probability regions, tends to work better with uniform quantizers (e.g., LSQ) that impose bounded activation ranges, thereby enabling more fine-grained modeling within a constrained domain. In contrast, JS divergence is symmetric and more tolerant to global distribution mismatch, making it a better match for non-uniform quantizers (e.g., N2UQ) whose activation statistics are inherently more flexible.

\subsubsection{Deployment Ablation: LayerNorm Removal Efficiency}
Due to space limitations, we move additional YOLO results to the \textbf{Appendix} \ref{Verification}, and present the MK-UNet results in the main text.
As summarized in Table~\ref{tab:ln_compact}, removing LayerNorm causes only a marginal average drop of 0.3\%, which is negligible for most practical applications, while a short post-folding fine-tuning stage recovers performance within minutes, indicating that the proposed folding procedure is practically efficient.


\begin{table}[!t]
\centering
\caption{LayerNorm removal on quantized MK-UNet (BUSI).}
\label{tab:ln_compact}
\small
\setlength{\tabcolsep}{1.0pt}
\resizebox{0.55\linewidth}{!}{%
\begin{tabular}{l|ccc}
\toprule
\textbf{Quantizer} & \textbf{With LN} & \textbf{Without LN} & Time(min) \\
\midrule
N2UQ & 60.7\% & 60.4\% & 5.6 \\
PACT & 46.3\% & 46.1\% & 3.4 \\
LSQ  & 49.7\% & 49.3\% & 3.6 \\
\bottomrule
\end{tabular}%
}
\end{table}

\section{Conclusion} 
In this work, we revisit low-bit quantization for complex vision models from a new perspective and identify a key factor behind the difficulty: biased gradient updates at feature fusion. To address this, we propose Q-GBFusion, a lightweight closed-loop fusion strategy that balances gradient flow across shallow and deep branches, and Q-ADA, a feature-aware supervision strategy that improves high-level semantic alignment during quantization-aware training. Extensive experiments demonstrate the effectiveness of both approaches.
\bibliographystyle{splncs04}
\bibliography{main}

\clearpage
\begin{center}
    {\LARGE \bfseries Appendix}
\end{center}
\vspace{0.5em}
\section{Visualization of Gradient Imbalance in Other Models}
\label{Imbalance}
\begin{figure}[!tp]
\centering
\begin{minipage}{0.45\linewidth}
\includegraphics[width=\linewidth]{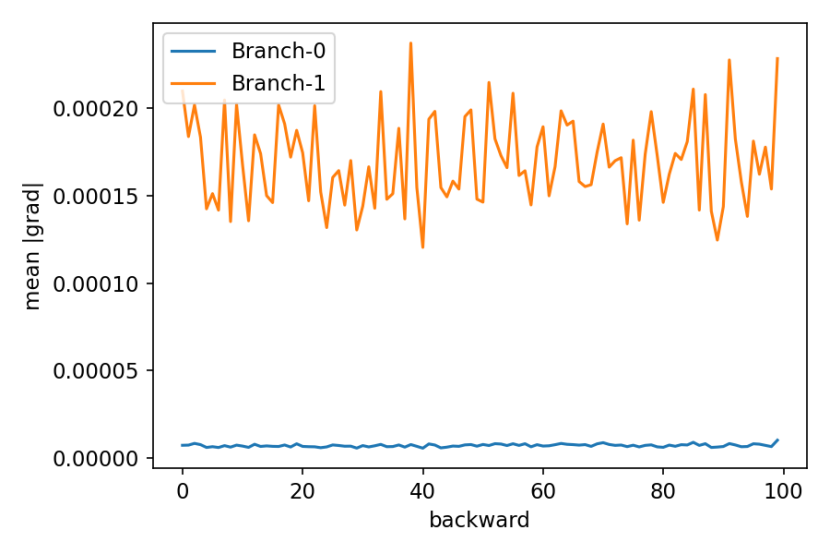}
\centerline{(a)}
\end{minipage}
\begin{minipage}{0.45\linewidth}
\includegraphics[width=\linewidth]{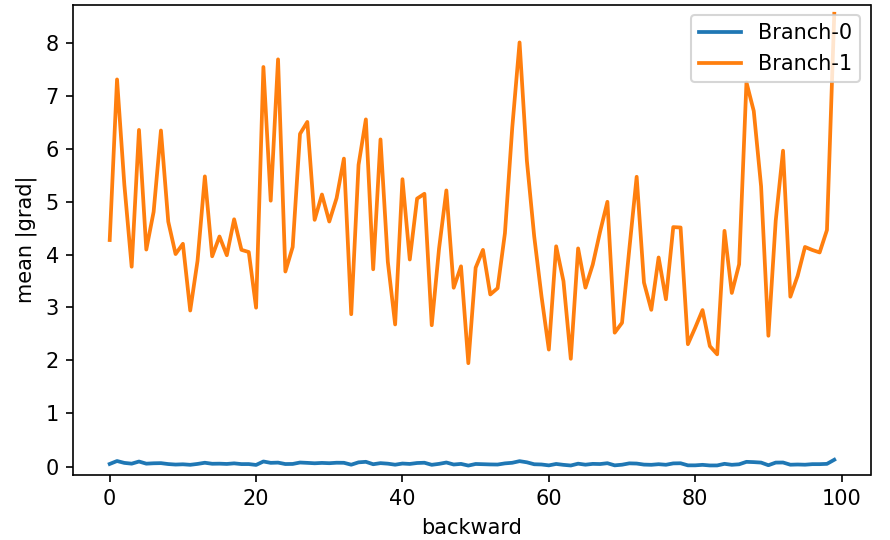}
\centerline{(b)}
\end{minipage}
\begin{minipage}{0.45\linewidth}
\includegraphics[width=\linewidth]{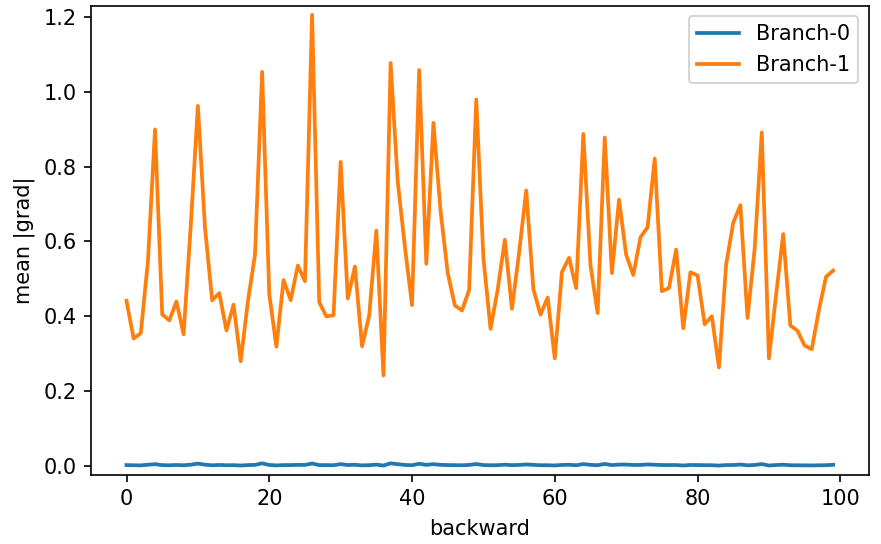}
\centerline{(c)}
\end{minipage}
\begin{minipage}{0.45\linewidth}
\includegraphics[width=\linewidth]{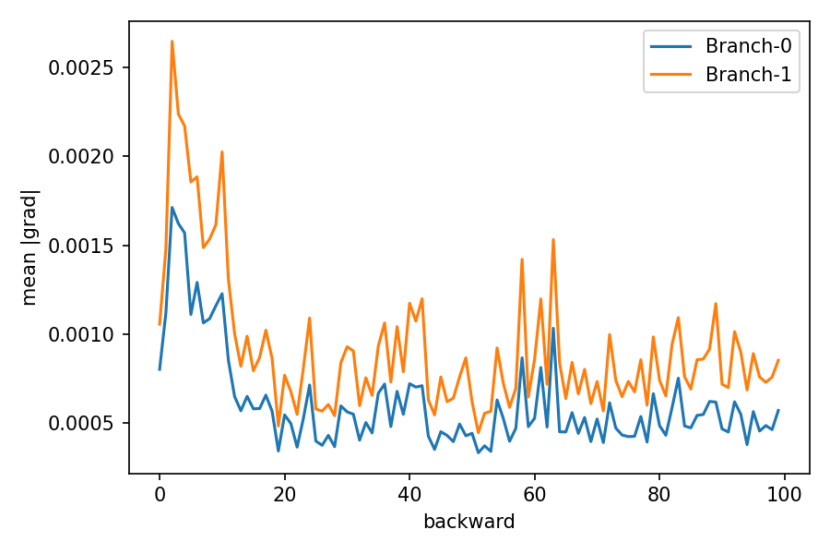}
\centerline{(d)}
\end{minipage}
\caption{(a) Gradient Measurement Results for YOLOv11. (b) Gradient Measurement Results for RT-DETR. (c) Gradient Measurement Results for MK-UNet. (d) Gradient Measurement Results for full-precision model(YOLOv11).}
\label{imbalance}
\end{figure}

Using the same tracking protocol as in the main text, we further trace and visualize the gradient curves of YOLOv11, RT-DETR, and MK-Unet, as shown in Fig.~\ref{imbalance}. The results show that gradient imbalance is not an isolated phenomenon specific to a single model, but a common behavior observed across different architectures. Furthermore, as shown in Fig.~\ref{imbalance}(d), this phenomenon is not evident in the full-precision model(YOLOv11), suggesting that it is closely related to the low-bit quantization process. These observations provide cross-model evidence for our motivation that complex vision networks share a structural bottleneck at feature fusion stages.

\section{How LayerNorm is Safely Removed}

\subsection{Theoretical Justification}
\label{LayerNormremove}

In quantized neural networks, LayerNorm removal is substantially easier than in full-precision networks. Because the dynamic range of fixed-point activations (e.g., 4-bit) is strictly constrained, the input statistics of LayerNorm exhibit much smaller fluctuations. As a result, LayerNorm can be accurately approximated by a fixed affine transformation estimated from calibration data, enabling safe and efficient removal without extensive retraining, in contrast to full-precision settings.

The LayerNorm layers introduced by our framework can be removed via the following steps:

 (1) After standard quantization-aware training (QAT) converges, we estimate fixed per-layer statistics  $(\mu_0,\sigma_0)$, i.e., the approximate mean and standard deviation of each LayerNorm input, using a small calibration dataset;

 (2) We then fold the resulting affine transformation into the adjacent convolutional layer as follows: 

Assuming that the standard layer normalization is defined by
\begin{equation}
    \mathrm{LN}(x) = \frac{x - \mu(x)}{\sqrt{\sigma^2(x) + \varepsilon}}
\end{equation}
where $\mu(x)$ and $\sigma^2(x)$ denote the mean and variance of the input. Then, the LayerNorm operation can be approximated as the following channel-wise affine transformation:
\begin{equation}
A = \frac{1}{\sqrt{\sigma_0^2 + \varepsilon}}, \qquad
B = - \frac{\mu_0}{\sqrt{\sigma_0^2 + \varepsilon}} .
\end{equation}
where $A$ and $B$ serve as per-channel scaling and bias terms that replace the LayerNorm operation during inference.




Assuming the network computes:
\begin{equation}
x \xrightarrow{\ \mathrm{LN}\ } \tilde{x} \xrightarrow{\ \mathrm{Conv}(W,b)\ } y ,
\end{equation}
where the LayerNorm output is an affine transform per channel: 
\begin{equation}
\tilde{x}_i = A_i x_i + B_i
\end{equation}
Let $W_{o,i,k,l}$ denote the convolution weight for output channel $o$, input channel $i$, and spatial position $(k,l)$, and $b_o$ the corresponding bias. Then, the convolution is then:
\begin{equation}
y_o
= \sum_{i,k,l} W_{o,i,k,l} \tilde{x}_{i,k,l} + b_o
= \sum_{i,k,l} W_{o,i,k,l} (A_i x_{i,k,l} + B_i) + b_o
\end{equation}
This shows that the LayerNorm affine parameters can be absorbed into the subsequent convolution, yielding equivalent parameters:

\begin{equation}
W'_{o,i,k,l} = W_{o,i,k,l} \cdot A_i ,
b'_o = b_o + \sum_{i,k,l} W_{o,i,k,l} B_i
\end{equation}
Replacing $(W, b)$ with $(W', b')$ leaves the forward pass unchanged, allowing the LayerNorm layer to be replaced by an identity mapping and safely removed during inference. 

(3) Finally, we perform post-LN-removal fine-tuning on a small subset of the quantization dataset for both tasks, using a unified strategy: no distillation is employed, and optimization relies solely on the native task losses. In both models, we use the Adam optimizer with a small learning rate of 2$\times10^{-5}$ to stably readjust feature distributions after LayerNorm removal.


\subsection{Experimental Verification}
\label{Verification}

Tables~\ref{ln1} compare the accuracy of the quantized YOLO models before and after LayerNorm removal. The results show that removing LayerNorm incurs only a marginal average degradation of 0.17\%, which is negligible for most practical applications. Moreover, the subsequent fine-tuning converges within minutes, demonstrating high efficiency.

\begin{table}[t]
\caption{Accuracy of the quantized YOLOv5 model on the PASCAL VOC dataset before and after LayerNorm removal. Reported time denotes fine-tuning duration.}
\label{ln1}
\centering
\begin{tabular}{lccc}
\toprule
Method & With LN & Without LN & Time(min) \\
\midrule
 N2UQ & 84.2\% & 84.1\% &  16.0 \\   
 PACT & 80.6\% & 80.6\% & 8.4 \\   
 LSQ & 78.9\% & 78.5\% &  9.0\\   
\bottomrule
\end{tabular}

\end{table}

\section{Final Experimental Setup}
\label{Setup}
All experiments were conducted on Ubuntu 20.04 with PyTorch 2.3.1 and CUDA 11.8. During training, we applied the Q-ADA strategy with a loss weight of 0.01, which may be slightly tuned depending on the quantization method.

\subsection{Object Detection Experimental Configuration}
For object detection on COCO and PASCAL VOC, we use SGD with models initialized from full-precision pretrained checkpoints. The initial learning rate is set to 0.00334, decayed via OneCycleLR with a final ratio of 0.15135. Momentum and weight decay are fixed at 0.74832 and 0.00025, respectively. All experiments use a batch size of 64, a fixed random seed of 0 for reproducibility, and 4 data-loader workers to balance I/O efficiency and system overhead.

\subsection{Image Segment Experimental Configuration}
For medical image segmentation on the BUSI dataset, we adopt Adam with an initial learning rate of $10^{-4}$, initializing from a full-precision MK-UNet checkpoint~\cite{Rahman_2025_ICCV}. UNet and its variants are canonical encoder–decoder architectures for segmentation, fusing shallow spatial details from the encoder with deep semantic features from the decoder. MK-UNet represents the latest advancement in this family and serves as our baseline.

\section{Reproduction Details}
\label{Reproduction}

To demonstrate the effectiveness of our framework, we compare it against several recent quantization-aware training (QAT) optimization methods, including QT-DoG~\cite{qtdog}, TR~\cite{2025scheduling}, HMQAT~\cite{huang2025hessian}, and EMA~\cite{EMA}. We also include EQ~\cite{ZHANG2024103277}, a recent post-training quantization (PTQ) method designed for quantizing segmentation models. Since these methods were not all originally developed for YOLO or MK-UNet, we reimplement them using the authors’ publicly released codebases and adapt them to our detection and segmentation benchmarks. Implementation details are as follows:

\textbf{QT-DoG\cite{qtdog}.}
For QT-DoG, we extract its core QAT mechanism, namely, the stage-wise quantization scheduling, and integrate it into our framework. Following the original protocol, we begin with 10 epochs of full-precision (FP32) training to stabilize the shared backbone, then introduce 4-bit activation quantization for the next 10 epochs to allow gradual adaptation to quantization perturbations across tasks, and finally enable full W4A4 quantization for the remaining epochs to achieve low-bit convergence. No modifications are made to the architecture, task heads, loss functions, or training pipeline, ensuring a strictly fair comparison.

\textbf{TR\cite{2025scheduling}.}
The TR scheduling mechanism, originally developed for classification, is adapted to our W4A4 setting by preserving its core principle: constraining the magnitude of latent weight updates to limit the transition frequency of quantized weights. We replace all learning rate based update logic in our framework with the TR driven update rule, effectively inserting a transition rate scheduler before the QAT optimizer. In this configuration, TR rather than the learning rate governs the update rhythm of quantized parameters, enforcing stable coarse to fine convergence. For fair comparison, the same TR schedule is applied to the shared backbone.

\textbf{HMQAT\cite{huang2025hessian}.}
We reformulate HMQAT’s Hessian-driven mixed-precision strategy within our  quantization framework as a second-order supervision mechanism. Rather than performing bit-width search, we leverage Hessian-based layer sensitivity estimates solely to guide gradient updates and quantization map adjustments. Following the original methodology, we compute the average Hessian trace for each layer using a full-precision teacher model, combine it with the layer’s parameter scale to derive sensitivity scores, and interpret these scores as priors for quantization tolerance: layers with high sensitivity are more vulnerable to quantization noise, whereas less sensitive layers can tolerate greater distortion. Importantly, we discard the mixed-precision search component entirely and retain only two core ideas—Hessian- and parameter-based sensitivity modeling, and sensitivity-aware fine-tuning during quantization-aware training.

\textbf{EMA\cite{EMA}.}
For EMA-based quantization, we reproduce its core contribution—smoothing of latent weights and quantization scale factors—while enforcing uniform W4A4 precision across all layers. During training, we maintain exponential moving average (EMA) versions of both backbone and task-head latent weights as well as quantization scales. We also implement its quantization correction (QC) mechanism, but without per-channel extensions or architectural modifications. In our setup, QC serves as a lightweight post-quantization calibration step: after completing W4A4 quantization-aware training, we freeze all task parameters and optimize a minimal set of affine correction factors on a small calibration set.

\textbf{EQ\cite{ZHANG2024103277}.}
To reproduce EfficientQ on MK-UNet, we begin with a pretrained full-precision model and apply layer-wise 8-bit post-training quantization. A single forward pass on one calibration sample is used to collect the full-precision output of each layer as the reconstruction target. For every convolutional layer, we first refine the activation quantization range via alternating minimization, then quantize the weights by solving a quadratic output-matching objective using ADMM, projecting the solution onto the discrete quantized set after each iteration. Following the original method, foreground regions are assigned higher reconstruction weights. The quantized output of each layer is propagated to the next, and no backpropagation or retraining is performed, resulting in an 8-bit post-training quantized version of MK-UNet for comparison.

\textbf{Compute-Optimal QAT\cite{dremov2026computeoptimal}.}
To reproduce this optimization strategy in our visual tasks, we adopt the paper’s cooldown--QAT fusion schedule while keeping the model, quantizer, losses, optimizer, data augmentation, and total training budget unchanged. Unlike the classic pipeline (full-precision training with cooldown, followed by QAT), we switch to QAT before the full-precision cooldown stage, enable fake quantization at the target bit-width, and perform the remaining learning-rate decay jointly with QAT (with a short QAT re-warmup after switching). 

\begin{figure*}
    \centering
    \includegraphics[width=0.99\linewidth]{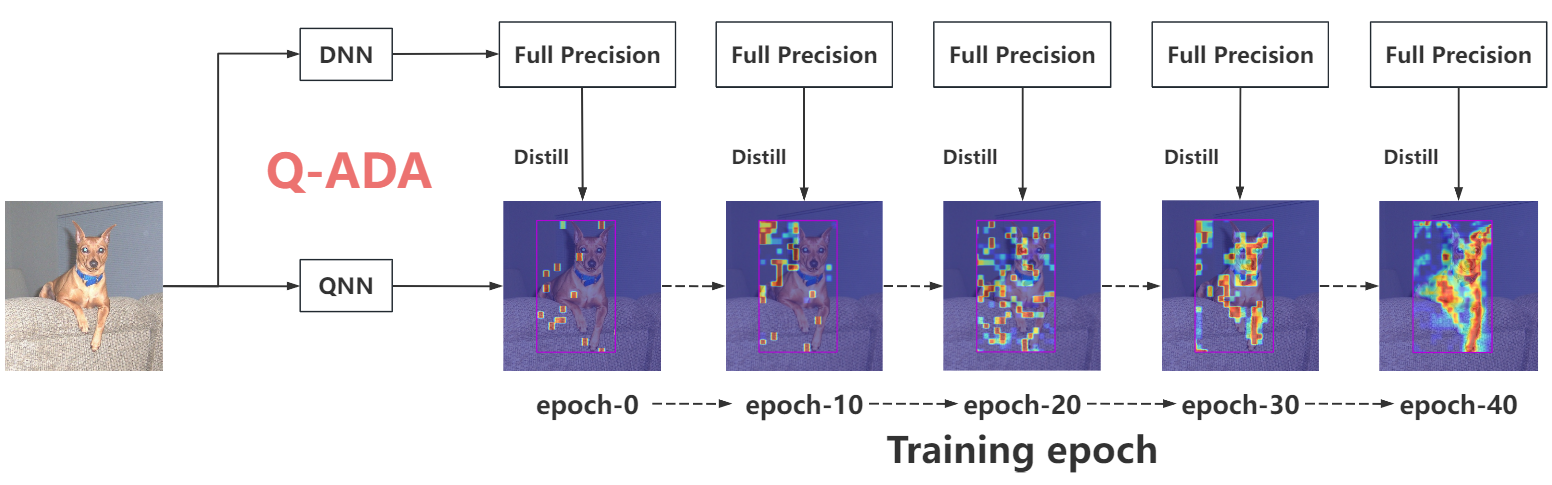}
    \caption{Workflow of Q-ADA. DNN denotes the full-precision baseline model, and QNN represents its quantized counterpart. Highlighted regions in the heatmaps indicate areas of critical semantic importance, thereby guiding the reduction of quantization errors during training.}
    \label{ADA}
\end{figure*}

\begin{table}[htbp]
\centering
\footnotesize
\caption{Supplementary experimental results of different quantizers on the COCO dataset.}
\label{tab:coco_quant}
\begin{tabular}{c c c c c}
\toprule
\textbf{Network} & \textbf{BW} & \textbf{Method} & $\mathbf{mAP_{50}}$ & $\mathbf{mAP}$ \\
\midrule
\multirow{6}{*}{\begin{tabular}[c]{@{}c@{}}YOLOv5s\end{tabular}}
& \multirow{6}{*}{W4A4}
& PACT    & 45.3\% & 28.3\% \\
& & PACT +Ours    & \textbf{47.0\%} & \textbf{29.2\%} \\
& & LSQ  & 44.8\% & 26.9\% \\
& & LSQ +Ours    & \textbf{45.8\%} & \textbf{27.9\%} \\
& & N2UQ   & 50.2\% & 31.1\% \\
& & N2UQ +Ours    & \textbf{51.4\%} & \textbf{33.2\%} \\
\midrule
\multirow{6}{*}{\begin{tabular}[c]{@{}c@{}}YOLOv11s\end{tabular}}
& \multirow{6}{*}{W4A4}
& PACT    & 51.2\% & 35.5\% \\
& & PACT +Ours    & \textbf{52.6\%} & \textbf{36.7\%} \\
& & LSQ  & 49.4\% & 33.0\% \\
& & LSQ +Ours    & \textbf{50.7\%} & \textbf{34.8\%} \\
& & N2UQ   & 57.0\% & 40.2\% \\
& & N2UQ +Ours    & \textbf{58.3\%} & \textbf{41.2\%} \\
\midrule
\multirow{6}{*}{\begin{tabular}[c]{@{}c@{}}RT-DETR\end{tabular}}
& \multirow{6}{*}{W4A4}
& Q-DETR   & 58.3\% & 40.7\% \\
& & Q-DETR +Ours & \textbf{59.6\%} & \textbf{41.9\%} \\
& & AQ-DETR & 59.0\% & 41.6\% \\
& & AQ-DETR +Ours & \textbf{60.6\%} & \textbf{43.0\%} \\
& & GPLQ    & 60.7\% & 43.8\% \\
& & GPLQ +Ours & \textbf{62.9\%} & \textbf{45.4\%} \\
\bottomrule
\end{tabular}
\end{table}



\section{Supplementary Experimental Results}

\subsection{Experimental Quantization Results on COCO}
\label{Supplementary Results}

Due to space constraints in the main paper, we reported only the results of combining N2UQ with our method on the COCO dataset, evaluated using the standard metric mAP\textsubscript{50–95}. In this section, we present additional results for multiple quantizers on COCO, including both mAP\textsubscript{50} and mAP\textsubscript{50–95} (denoted as mAP). As shown in Table~\ref{tab:coco_quant}, experiments on the COCO benchmark also demonstrate that our strategy consistently improves performance by 1\% to 2\% in both mAP\textsubscript{50} and mAP across a range of mainstream quantization baselines, highlighting its effectiveness and broad applicability.

\subsection{Additional Empirical Evidence}
\label{addition}

To further disentangle whether the low-bit degradation can be mitigated by simple optimization tuning, we run a controlled ablation that applies \emph{fixed} branch-wise learning-coefficient scaling between the shallow and deep branches, and report the results in Table~\ref{tab2}. Overall, varying a constant ratio yields \emph{non-monotonic} and \emph{inconsistent} changes in accuracy (e.g., a mild gain at $\times$4 but a clear drop at $\times$8), indicating that the observed degradation cannot be reliably resolved by a static re-weighting schedule. This supports our hypothesis that low-bit quantization induces a \emph{structured} learning imbalance at feature-fusion modules, which calls for \emph{adaptive} rebalancing rather than hand-tuned fixed coefficients. 

In addition, when applied to the full-precision model, Q-GBFusion does not improve performance (and remains within normal training variance), suggesting that the module primarily acts as a corrective mechanism under quantization, rather than a generic architectural enhancement.

\begin{table}[htbp]
\centering
\small
\caption{Negative control: fixed branch-wise coefficient scaling is insufficient (\,$\times N$: shallow/deep = $N$\,)}
\label{tab2}
\resizebox{0.85\linewidth}{!}{
\begin{tabular}{c c c c c }
\toprule
\textbf{Coeff. Ratio} & \textbf{BW} & \textbf{Network} & \textbf{Quantizer} & \textbf{mDICE} \\
\midrule
$\times$1 & W4A4  & MK-Unet & N2UQ & 55.4\% \\
$\times$2 & W4A4  & MK-Unet & N2UQ & 55.6\% \\
$\times$4 & W4A4  & MK-Unet & N2UQ & 55.9\% \\
$\times$8 & W4A4  & MK-Unet & N2UQ & 53.8\% \\
\midrule
$\times$1 & FP32  & MK-Unet & -- & 69.5\% \\
$\times$1 & FP32  & MK-Unet + Q-GBFusion & -- & 69.3\% \\
\bottomrule
\end{tabular}
}
\end{table}

\section{Visualizing the Mechanism of Q-ADA}
\label{Visualizing}

As shown in Fig.~\ref{ADA}, during the early stages of quantization training, the substantial quantization error introduced by naive quantization severely disrupts critical attention patterns. This results in significant degradation of semantic perception, particularly within target regions, and manifests as pronounced feature attenuation and information collapse. In effect, the model at this stage suffers from both severe semantic information loss and high levels of quantization noise.

As the distillation process progresses, soft supervision from the teacher model gradually guides the quantized model to realign its attention distribution. Responses associated with spatial focus, target shape, and boundary regions are progressively restored. Meanwhile, quantization error diminishes significantly over the course of training, enabling the model to locate and represent target regions more accurately and ultimately yielding stable and reliable quantized feature representations.

\end{document}